\documentclass[lettersize,journal]{IEEEtran}
\usepackage{xcolor}
\usepackage{algorithm}
\usepackage{arydshln}
\usepackage{algorithmic}
\usepackage{threeparttable}
\usepackage{xtab}
\usepackage{array}
\usepackage{multirow}
\usepackage{booktabs}
\usepackage{amsmath}
\usepackage{colortbl}
\usepackage{cite}
\usepackage[caption=false,font=normalsize,labelfont=sf,textfont=sf]{subfig}
\usepackage{textcomp}
\usepackage{stfloats}
\usepackage{amsfonts}
\usepackage{verbatim}
\usepackage{graphicx}
\usepackage{wasysym}

\begin{document}
\title{Defensive Adversarial CAPTCHA: A Semantics-Driven Framework for Natural Adversarial Example Generation}
\author{Xia Du, Xiaoyuan Liu,  
 Jizhe Zhou$^*$, Zheng Lin, Chi-man Pun,~\IEEEmembership{Senior Member,~IEEE}, Cong Wu, Tao Li, Zhe Chen,~\IEEEmembership{Member,~IEEE}, Wei Ni,~\IEEEmembership{Fellow,~IEEE}, Jun Luo,~\IEEEmembership{Fellow,~IEEE}

\thanks{Xiaoyuan Liu and Xia Du are with the School of Computer and Information Engineering, Xiamen University of Technology, Xiamen, 361000, China (email: 2322071027@stu.xmut.edu.cn; duxia@xmut.edu.cn; ).}
\thanks{Jizhe Zhou is with the School of Computer Science, Engineering Research Center of Machine Learning and Industry Intelligence, Sichuan University, Chengdu, China, 610020, China (email: yb87409@um.edu.mo).}
\thanks{Zheng Lin, Cong Wu, and Tao Li are with the Department of Electrical and Electronic Engineering, University of Hong Kong, Pok Fu Lam, Hong Kong, China (e-mail: linzheng@eee.hku.hk; congwu@hku.hk; lthku999@connect.hku.hk).}
\thanks{Chi-man Pun is with the Department of Computer and Information Science, Faculty of Science and Technology, University of Macau, Macau, 999078, China (email: cmpun@umac.mo).} 
\thanks{Zhe Chen is with the Institute of Space Internet, Fudan University, Shanghai 200438, China, and the School of Computer Science, Fudan University, Shanghai 200438, China (e-mail: zhechen@fudan.edu.cn).}
 \thanks{Wei Ni is with Data61, CSIRO, Marsfield, NSW 2122, Australia, and the School of Computing Science and Engineering, and the University of New South Wales, Kensington, NSW 2052, Australia (e-mail:
wei.ni@ieee.org).}
\thanks{Jun Luo is with the School of Computer Engineering, Nanyang Technological University, Singapore (e-mail: junluo@ntu.edu.sg).}
\thanks{$^*$ denotes Corresponding author.}
\thanks{
Corresponding author: Jizhe Zhou (yb87409@um.edu.mo)
}
}

\markboth{Journal of \LaTeX\ Class Files,~Vol.~14, No.~8, August~2021}%
{Shell \MakeLowercase{\textit{et al.}}: A example Article Using IEEEtran.cls for IEEE Journals}


\maketitle

\begin{abstract}

Traditional CAPTCHA (Completely Automated Public Turing Test to Tell Computers and Humans Apart) schemes are increasingly vulnerable to automated attacks powered by deep neural networks (DNNs). Existing adversarial attack methods often rely on the original image characteristics, resulting in distortions that hinder human interpretation and limit their applicability in scenarios where no initial input images are available. To address these challenges, we propose the Unsourced Adversarial CAPTCHA (DAC), a novel framework that generates high-fidelity adversarial examples guided by attacker-specified semantics information. Leveraging a Large Language Model (LLM), DAC enhances CAPTCHA diversity and enriches the semantic information. To address various application scenarios, we examine the white-box targeted attack scenario and the black-box untargeted attack scenario. For target attacks, we introduce two latent noise variables that are alternately guided in the diffusion step to achieve robust inversion. The synergy between gradient guidance and latent variable optimization achieved in this way ensures that the generated adversarial examples not only accurately align with the target conditions but also achieve optimal performance in terms of distributional consistency and attack effectiveness. In untargeted attacks, especially for black-box scenarios, we introduce bi-path unsourced adversarial CAPTCHA (BP-DAC), a two-step optimization strategy employing multimodal gradients and bi-path optimization for efficient misclassification. Experiments show that the defensive adversarial CAPTCHA generated by BP-DAC is able to defend against most of the unknown models, and the generated CAPTCHA is indistinguishable to both humans and DNNs.

\end{abstract}


\section{Introduction}
\label{sec:intro}

\IEEEPARstart{C}{APTCHA} (Completely Automated Public Turing Test to Tell Computers and Humans Apart) is a fundamental cybersecurity mechanism. Its core function is to distinguish legitimate human users from automated bots. This technology presents specific computational challenges that are easily solvable by humans but difficult for machines. By filtering out automated attacks, these mechanisms safeguard digital services while maintaining system integrity.


\begin{figure}
    \centering
    \includegraphics[width=1\linewidth]{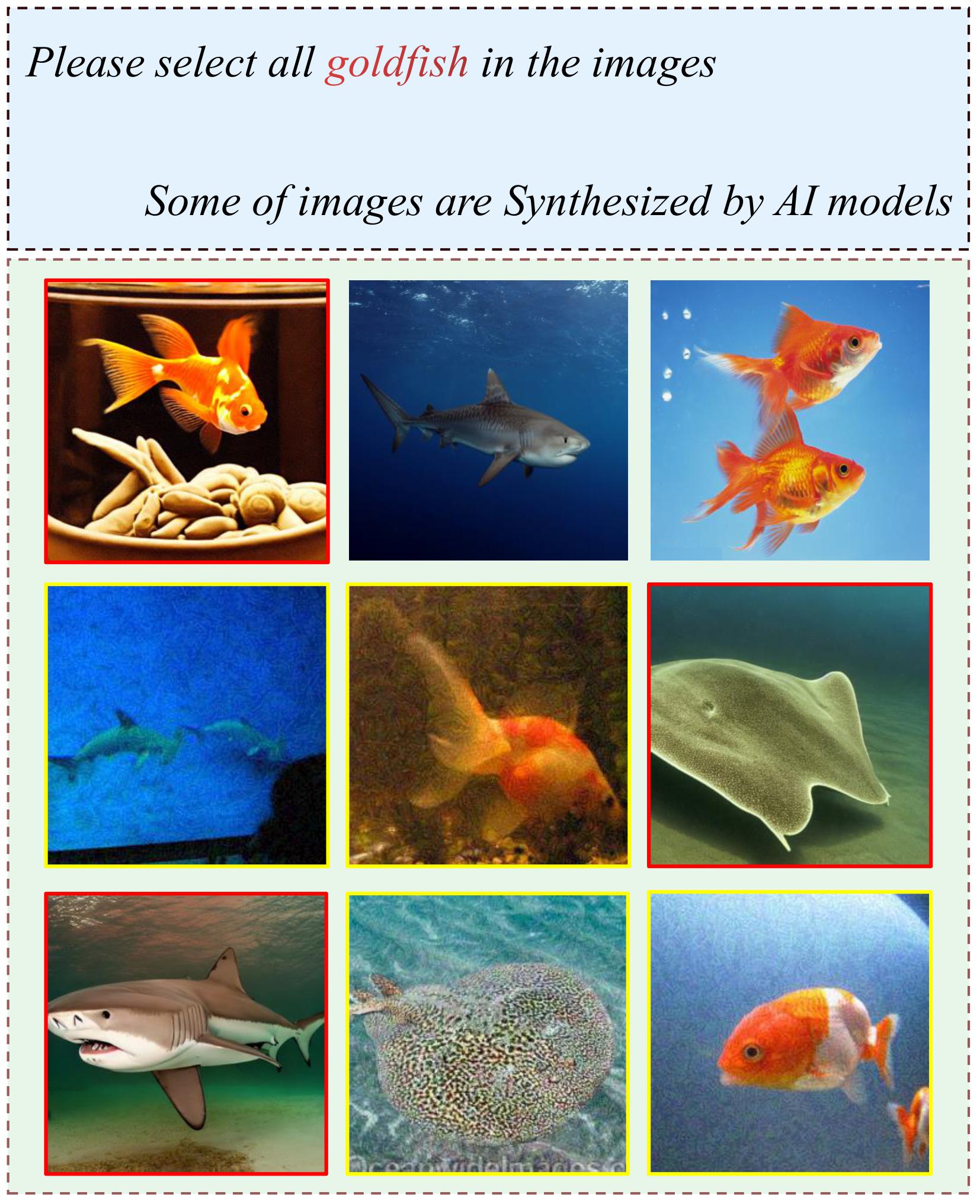}
    \caption{Practical application scenarios of adversarial examples in CAPTCHA images. To allow readers can distinguish the source of the images, we label the adversarial examples generated by our method with red boxes, the adversarial examples generated by the traditional method with yellow boxes, and the clean images that are not labeled.}
    \label{fig:enter-label}
\end{figure}

Generally speaking, the effectiveness of a CAPTCHA can be evaluated based on its security. This refers to the CAPTCHA's ability to defend against unknown hacking techniques used by attackers, as well as its user-friendliness. However, with the rapid development of artificial intelligence~\cite{lin2024efficient,fang2024automated,yuan2024satsense,hu2024agentscodriver,peng2024sums,lin2024adaptsfl,lin2025hsplitlora}, DNN models have achieved remarkable results in fields such as image recognition and image segmentation~\cite{fang2024ic3m,lin2025leo,yuan2025constructing,lyu2023optimal,hu2024accelerating,lin2024hierarchical}. They also pose unintended challenges in areas where maintaining human control and security is critical. For instance, image recognition technologies based on DNN models have grown highly effective at deciphering even the most complex CAPTCHA systems. CAPTCHAs, originally designed to differentiate between humans and machines, are now increasingly vulnerable to attacks powered by advanced deep learning models. These models can break CAPTCHA defenses with remarkable precision, rendering conventional CAPTCHA systems less effective at safeguarding online platforms against automated attacks.

\begin{figure*}
    \centering
    \includegraphics[width=1\linewidth]{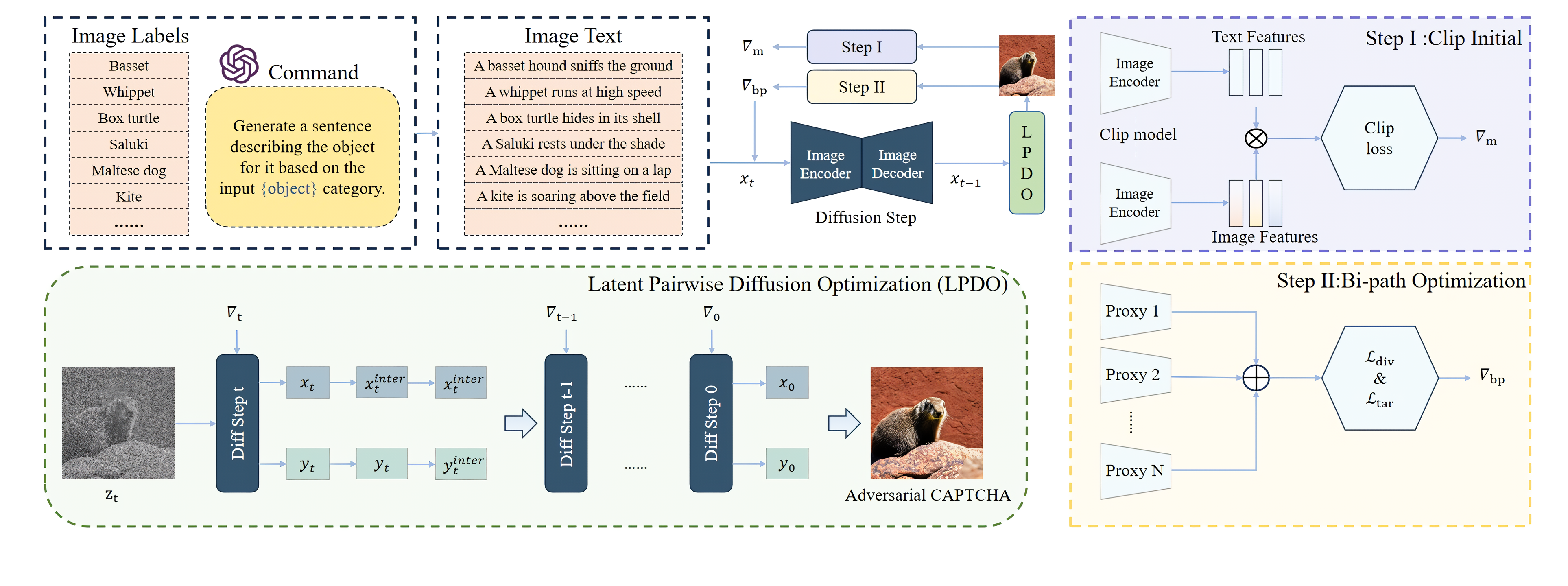}
    \caption{Our proposed bi-path unsourced adversarial CAPTCHA (BP-DAC) attack framework.}
    \label{fig:frame}
\end{figure*}

Adversarial attacks~\cite{zhang2025robust,duan2025rethinking,zhang2025state}, which exploit the sensitivity of deep learning models to specific and carefully designed perturbations, have emerged as a powerful tool to evaluate model robustness \cite{gong2017crafting,cisse2017houdini}. These attacks introduce subtle, targeted noise to mislead models into incorrect predictions, serving as an effective tool for evaluating the robustness of AI systems. This concept has been creatively extended to adversarial CAPTCHAs, where adversarial perturbations are used to enhance CAPTCHA robustness against automated recognition. This novel approach has started to gain traction in recent years, particularly among leading AI companies, and represents an innovative change in security strategies for CAPTCHA systems \cite{7954632,7467367}.

Although adversarial CAPTCHAs represent a significant advancement in system security against automated attacks, adversarial CAPTCHA systems still face numerous challenges. The limited availability of image types (due to copyright restrictions, etc.) results in insufficient diversity of adversarial samples. Additionally, while increasing perturbation intensity to ensure strong defense against black-box models, this approach compromises visual naturalness, thereby harming user experience and usability. Furthermore, generating CAPTCHAs that are both natural and widely resistant to DNN attacks requires a vast search space, which traditional pixel-level perturbation methods struggle to handle. Diffusion models offer a new direction for enhancing both security and usability, as their potential to unlock search space enables the generation of more natural yet adversarial images, thereby strengthening robustness. However, existing diffusion model-based attack techniques still rely on original samples, limiting the full realization of their adversarial strength.

To address these issues, we designed the DAC framework. By shifting input information from the pixel-level to the semantic level, we overcame the limitations of traditional adversarial sample generation methods, which are constrained by the pixel search space. This enables defensive CAPTCHA designers to generate adversarial CAPTCHA images rich in semantic information simply by inputting the desired image categories. This reduces the need for data collection and manual annotation of verification images. Specifically, before the attack, in the face of white-box scenarios that are easier to defend against, we utilize LLM to convert the categories entered by CAPTCHA designers into more complex semantic information, thereby ensuring the accuracy of the generated images. Additionally, unlike other adversarial attack methods based on diffusion models, we design parameter-shared coupled noise vector pairs during the diffusion model generation process, achieving robust inversion through an iterative bidirectional calibration mechanism at discrete time steps. This method approximates the initial input by performing a denoising operation at each time step, allowing the model gradient to propagate backward through the entire chain. This solves the inherent instability and error accumulation problems of diffusion-based adversarial attack methods. It ensures the stability and high quality of adversarial example generation. In more challenging and practical black-box attack scenarios, we further propose the BP-DAC framework, which proposes a novel bi-path optimization strategy. This method enhances optimization efficiency and attack performance by further aggregating multi-model gradients during the attack optimization process while simultaneously guiding the generative model in the direction of the suboptimal and lowest category gradients of the target model.

In summary, our contributions are as follows:

$\bullet$ To address the current issues with CAPTCHA systems, we propose the DAC framework, which innovatively transforms the input of CAPTCHA creators from clean existing images to semantic information. By incorporating EDICT into the process of generating adversarial samples using diffusion models, our method achieves high-fidelity adversarial sample generation through precise coupling inversion. Compared to existing methods, it generates more accurate images while reducing error accumulation, maintaining stealthiness, and without requiring additional training or data adaptation.

$\bullet$ When faced with unknown black-box models, which are more common in real-world applications, we further proposed the BP-DAC framework, which integrates the gradients of multiple models into the guidance process and simultaneously guides the attack toward the suboptimal category and lowest category gradient of the target model, thereby improving optimization efficiency and the defense capability of CAPTCHAs against unknown models.

$\bullet$ Extensive experiments have demonstrated that our approach can guarantee a high attack success rate (ASR) for deep neural network models with different structures under different white- and black-box scenarios while generating unsourced Adversarial CAPTCHAs that are indistinguishable from clean examples.

To the best of our knowledge, we are the first to generate an unsourced adversarial CAPTCHA using the diffusion model. At the same time, by means of a novel bi-path optimization strategy, we achieve for the first time a near 99.4\% ASR against an unknown black-box model when generating adversarial examples using the generative model. This opens up new possibilities for adversarial research.

\section{RELATED WORK}
As the domain of adversarial attacks continues to mature, the endeavor to render adversarial perturbations increasingly inconspicuous has emerged as a pivotal topic.

In recent years, various innovative approaches to adversarial image attacks have emerged. MUTEN \cite{guo2021mutenboostinggradientbasedadversarial} enhances the success rate and robustness of gradient-based adversarial attacks by utilizing diverse variant models. GADT \cite{ma2024gadtenhancingtransferableadversarial} improves the migrability of adversarial examples by optimizing the data enhancement parameters, which is particularly suitable for black-box and query attacks. MGAA \cite{9711246} improves the mobility of attacks using meta-learning methods and enhances the success rate of attacks by narrowing the difference in gradient directions in white-box and black-box environments. U-GAN \cite{song2018constructingunrestrictedadversarialexamples} constructs unconstrained adversarial examples through GAN networks, breaking through the restriction on the perturbation range of traditional gradient-based attack methods, making the generated adversarial example more aggressive and effective in bypassing many existing defense mechanisms. AdvDiff \cite{dai2024advdiffgeneratingunrestrictedadversarial} generates unconstrained adversarial examples by utilizing the denoising process of the diffusion model, demonstrating the potential of this model in adversarial attacks. DiffAttack \cite{kang2024diffattackevasionattacksdiffusionbased} proposes an attack strategy specifically for countering the purification defense of the diffusion model, which successfully generates adversarial examples by bypassing the purification process, revealing potential loopholes in the defense mechanisms of the diffusion model and driving new challenges in the field of adversarial attacks. 

Not only in the field of image recognition, but also in the field of security, researchers have attempted to incorporate undetectable perturbations into CAPTCHAs to counter machine intrusion. Shi \textit{et al.} \cite{9440853} proposed an aCAPTCHA system, which enhances the security of ordinary CAPTCHAs by generating adversarial examples. This approach makes it difficult for deep learning models to recognize them by adding adversarial perturbations to the images, while still allowing human users to pass normally. Wen \textit{et al.} \cite{Wen2023RobustIC} explored methods to generate stronger CAPTCHA challenges through adversarial attacks, such as Iterative FGSM (I-FGSM) \cite{kurakin2017adversarialexamplesphysicalworld} and DeepFool \cite{moosavidezfooli2016deepfoolsimpleaccuratemethod}. Their method, which focuses on perturbation processing, enhances the resistance of these visual challenges to machine learning models, making them more difficult for automated systems to crack. Zhang \textit{et al.} \cite{8644665} explored the application of adversarial examples on different image classes of CAPTCHAs, and investigated how to improve the robustness of image CAPTCHAs by using adversarial methods and analyzing the effect of adversarial example perturbation on it.

Although the above methods can generate natural adversarial examples using various techniques, they still require corresponding benign images to generate adversarial examples, or they cannot generate adversarial examples with complex semantic information. This is because the fundamental flaw of adversarial attack methods in this regard limits the search space of perturbations, and the high intensity of attack performance may lead to excessive perturbations, thereby affecting human visual perception. Unlike the above work, our approach breaks out of the traditional pixel space and generates complex adversarial examples based on intricate semantic information.

\begin{table}[h]
\caption{{Characteristics of adversarial attack methods. \Circle/\RIGHTcircle/\CIRCLE indicate the low/middle/high quality of the methods.}}
\label{tab:metrics}
\resizebox{\linewidth}{!}{
\centering
\begin{tabular}{c|ccccc}
\toprule[0.15em]
\multirow{2}{*}{Attack Methods} & \multicolumn{4}{c}{  Method Characteristics } \\ \cmidrule(l){2-5}
 & {  Dependent input    }  & {   Diversity  } & {  Quality   } &{ transferability   }  \\ \midrule
MUTEN \cite{guo2021mutenboostinggradientbasedadversarial}  & \Circle & \Circle & \RIGHTcircle &  \Circle   \\  
GADT \cite{ma2024gadtenhancingtransferableadversarial}  & \Circle& \Circle & \RIGHTcircle &   \RIGHTcircle  \\  
MGAA \cite{9711246} & \Circle& \Circle & \RIGHTcircle & \CIRCLE  \\

U-GAN \cite{xiao2019generatingadversarialexamplesadversarial} & \Circle & \RIGHTcircle &  \CIRCLE &\Circle   \\
DiffAttack \cite{kang2024diffattackevasionattacksdiffusionbased}  & \Circle & \RIGHTcircle &  \CIRCLE   &  \Circle  \\      

AdvDiff \cite{dai2024advdiffgeneratingunrestrictedadversarial}  & \Circle & \RIGHTcircle &  \CIRCLE   &  \RIGHTcircle  \\   


\midrule
DAC (ours)    & \CIRCLE & \CIRCLE  & \CIRCLE  & \Circle   \\ 
BP-DAC  (ours)     & \CIRCLE & \CIRCLE & \CIRCLE   & \CIRCLE   \\

\midrule [0.15em]
\end{tabular}
}
 
\label{tab:pitch}
\end{table}


\section{BACKGROUND}
\label{sec:work}

\subsection{Traditional adversarial scenarios}
Traditional adversarial attacks can be categorized into untargeted and targeted attacks. To cope with different application scenarios, our approach discusses both targeted and untargeted attacks. Targeted attacks aim to shift the model's predictions to a targeted category specified by the attacker. This attack can be considered “targeted” because the attacker wants the model to output a specific mislabel when confronted with a modified example. Specifically, a traditional targeted attack is formulated as follows:
\begin{equation}
\begin{gathered}
   x^{\prime}=\arg\min_{x^{\prime}}\ell(f(x^{\prime}),y_{\mathrm{target}}) \\
\begin{aligned}
    \mathrm{s.t.}\: x^{\prime}=x+\delta. 
\end{aligned}
\end{gathered}
\label{eq:1}
\end{equation}

Instead, the untargeted attack aims to make the prediction results different from the original category without specifying the wrong category. The attacker only needs the model to output any of the wrong categories when it sees the antagonistic example:
\begin{equation}
\begin{gathered}
   x^{\prime}=\arg\max_{x^{\prime}}\ell(f(x^{\prime}),y_{\mathrm{target}}) \\
\begin{aligned}
    \mathrm{s.t.}\: x^{\prime}=x+\delta  .
\end{aligned}
\end{gathered}
\label{eq:2}
\end{equation}

From Eq.(\ref{eq:1}) and Eq.(\ref{eq:2}), it can be concluded that \textbf{traditional adversarial attacks rely on specify benign images $x$ as the basis for generating adversarial examples $x^{\prime}$.}

\subsection{Diffusion model}

Denoising Diffusion Models (DDMs) \cite{nichol2021improveddenoisingdiffusionprobabilistic} are a type of generative model that produces images by progressively suppressing noise, starting from pure noise and iterating until a clean image is generated. The process consists of two stages: Forward diffusion and reverse diffusion.

Forward Diffusion Process: In this stage, clean images are corrupted by adding Gaussian noise in a series of steps, eventually transforming the image into pure noise. At each step $t$, the noisy image $x_t$ is computed as 
\begin{align}
     x_ {t+1}  =  \sqrt {a_ {t}x_t+\sqrt {1-\alpha _ {t}\epsilon  }}, 
\end{align}
where $a_ {t}$ is the noise-scaling factor, and $\epsilon$ represents the Gaussian noise.

Reverse Diffusion Process: Starting from pure noise, the model iteratively denoises the image, generating a series of intermediate images that gradually resemble the original image. The reverse process is modeled as
\begin{align}
p_{\theta}\left(x_{t-1} \mid x_{t}\right) & = \mathcal{N}\left(x_{t-1} ; \mu_{\theta}\left(x_{t}, t\right), \Sigma_{\theta}\left(x_{t}, t\right)\right),
\end{align}
where $\mu_{\theta}$ and $\Sigma_{\theta}$ are the learned parameters for mean and variance,  respectively, guiding the denoising process.

To accelerate sampling, Denoising Diffusion Implicit Models (DDIM) \cite{song2022denoisingdiffusionimplicitmodels} introduce a non-stochastic, efficient sampling approach. DDIM avoids fully stochastic sampling, offering a faster alternative by approximating the reverse steps as
\begin{align}
    x_{t-1}=\sqrt{\bar{\alpha}_{t-1}}\cdot\mu_\theta(x_t,t)+\sqrt{1-\bar{\alpha}_{t-1}}\cdot\epsilon,
\end{align}
which reduces the number of steps required to generate an example, making the model more practical for real-time applications.



\subsection{EDICT}
In recent years, EDICT \cite{wallace2022edictexactdiffusioninversion} has been proposed to address the problem of reconstruction and content loss in model-generated images due to error propagation during DDM backpropagation \cite{rombach2022highresolutionimagesynthesislatent}. EDICT mathematically and precisely inverts real- and model-generated images by coupling latent noise variables. Specifically, EDICT utilizes two alternating sequences, $x_t$ and $y_t$, in the symmetric coupling layer of the DDM. These sequences are used to track and reconstruct the states of latent variables during the generation process. The forward updating of $x_t$ and $y_t$ during the DDM inverse propagation is then realized through the following coupling steps:
\begin{equation}
    \begin{aligned}
x_{t}^{inter}& =a_t\cdot x_t+b_t\cdot\Theta_{(t,C)}(y_t) ;\\
y_{t}^{inter}& =a_t\cdot y_t+b_t\cdot\Theta_{(t,C)}(x_t^{inter}); \\
x_{t-1}& =p\cdot x_{t}^{inter}+(1-p)\cdot y_{t}^{inter}; \\
y_{t-1}& =p\cdot y_{t}^{inter}+(1-p)\cdot x_{t-1} ,
\end{aligned}
\end{equation}
where $a_t$ and $b_t$ are time-related coefficients, EDICT includes the internal transformations and backward updates of the variables. It generates a new $x_{t-1}$ and $y_{t-1}$ by mixing the internal representations of $x_t$ and $y_t$ through the averaging parameter $p\in (0,1)$, which is derived using linear equations through the above equation to ensure that the entire diffusion process can be performed exactly in the reverse order:
\begin{equation}
    \begin{aligned}
y_{t+1}^{inter}& =(y_{t}-(1-p)\cdot x_{t})/p \\
x_{t+1}^{inter}& =(x_t-(1-p)\cdot y_{t+1}^{inter})/p \\
y_{t+1}& =(y_{t+1}^{inter}-b_{t+1}\cdot\Theta_{(t+1,C)}(x_{t+1}^{inter}))/a_{t+1} \\
x_{t+1}& =(x_{t+1}^{inter}-b_{t+1}\cdot\Theta_{(t+1,C)}(y_{t+1}))/a_{t+1}.
\end{aligned}
\end{equation}

\begin{figure}[t]
\centering
  \includegraphics[width=1\linewidth]{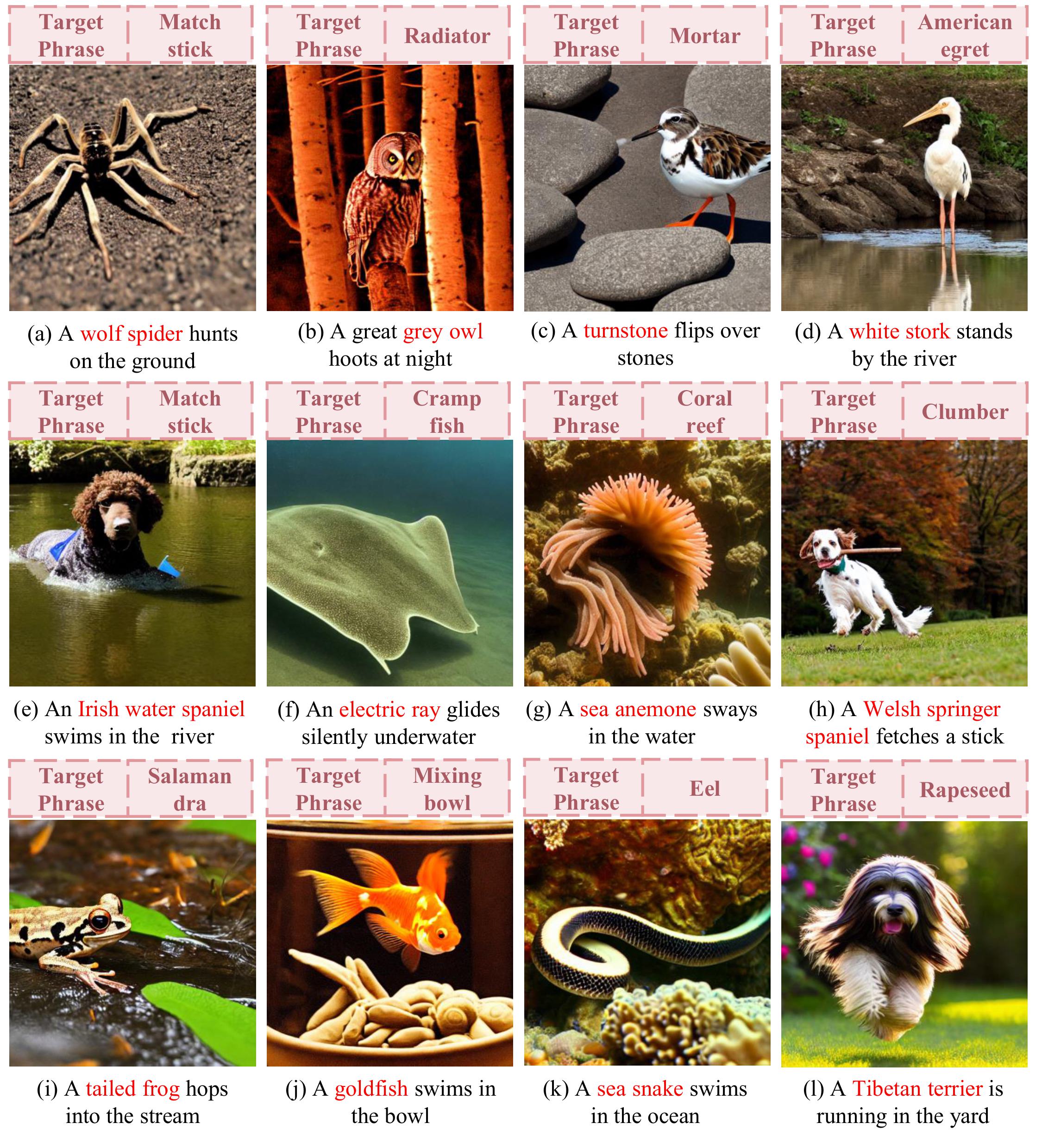}
  \caption{The adversarial examples generated by our proposed method.}
  \label{compare}
\end{figure}

\section{METHODOLOGY}
\label{section2}




In this section, we introduce the DAC framework in the context of white-box attacks, where we integrate both the LLM and EDICT frameworks to generate unsourced adversarial examples while ensuring the quality of the generated images. To address the broader applicability and higher complexity of black-box attack scenarios, we further extend the DAC framework by incorporating the clip model and the bi-path optimization strategy. This enhancement ensures semantic consistency during the diffusion model's generation process and overcomes the limitations of traditional search spaces, enabling the generation of unsourced adversarial CAPTCHAs suitable for practical application scenarios. The overall framework is shown in Fig. \ref{fig:frame}.

\subsection{Threat model}

Considering the distinct contexts of image recognition and CAPTCHA security, we adopt a threat model based on image classification and examine both white-box and black-box attack scenarios. In this scenario, the attacker only needs to provide the target class and the desired adversarial instance class during the generation phase, and manipulate the test image during the inference phase without involvement in the training phase of the model.

\subsection{Defensive adversarial CAPTCHA}
\subsubsection{Prompt guidance}

First, we introduce an LLM to ensure the accuracy and diversity of the generation process. Although the initial attack aims to generate corresponding adversarial examples based on the object categories entered by the attacker, our experiments reveal that using only short prompts $P^{\prime}$ tends to lead to generation errors. This is because brief prompts lack sufficient contextual information, making the diffusion model ambiguous in understanding the target image features. Diffusion models rely on the semantics of the prompter to gradually guide image generation. When the prompt is too simple, for example, a single word or an ambiguous description, the model may associate it with multiple different potential image features, resulting in generation results that deviate from expectations.

To address this issue, we introduce an LLM \cite{touvron2023llamaopenefficientfoundation,vicuna2023, TheC3} to enhance the semantic information of the prompt. The LLM guides the generator's inputs by generating richer and contextually relevant prompts like ``A goldfish swims in the bowl", which not only ensures the diversity and stability of the generated examples but also enables the attacker to generate more accurate adversarial examples, given the category and the specific target. That can be written as:
\begin{align}
    P^{\prime}=f_{\mathrm{LLM}}(P).
\end{align}

Next, we utilize a diffusion model $G$ to generate images step by step, with the generation process of this model conditionally guided by the extended prompt. Specifically, we first input the extended prompt $P^{\prime}$ into $G$ for conditional generation, and $G$ generates a representation of the intermediate latent variable $z_t$ based on the latent variables and the prompt at each time step $t$. Specifically, this process can be formulated as: 
\begin{align}
    I_{t}=G(z_t,P^{\prime},\epsilon_t),
\end{align}
where $\epsilon_t$ is the noise term that we introduce in the generation process $t$ to ensure diversity in the generation.



\begin{algorithm}[t]
\caption{Adversarial example synthesis in DAC}
\label{alg:adversarial example synthesis}
\begin{algorithmic}[1]
\REQUIRE Initial prompt $P$ from the attacker, generator $G$, LLM $f_{LLM}$, target category $y_{target}$, iterations $T$, learning rates $\eta_x$, $\eta_y$, weight $\alpha$, cross-entropy loss $L$
\ENSURE $f(I_t) = y_{target}$;
\STATE $P' = f_{LLM}(P)$;
\STATE Initialize latent variables $x$ and $y$
\FOR{$t = T$ to $1$}
\STATE $I_{t}=G(z_t,P^{\prime},\epsilon_t)$;
\IF {$f(I_t)$ = $y_{target}$}
\STATE exit the loop and proceed to Step 17
\ELSE
\STATE Introduce gradients $\nabla$ of $f$
\STATE Compute loss $L$ between generated image $I_t$ and $y_{target}$
\STATE Update latent variables $x$ and $y$:\\
\STATE$x^{inter}_{t-1} = x_t - \eta_{x} * \nabla_{x_t} \ell(x_t,y_{target})$,
\STATE$y^{inter}_{t-1} = y_t - \eta_y * \nabla_{y_t}\ell(y_t,y_{target})$;
\STATE Merge optimized latent variables into a unified representation:\\
$z_{t-1} = \alpha * x^{inter}_{t-1} + (1 - \alpha) * y^{inter}_{t-1}$; 
\STATE $I_{t-1} = G(z_{t-1})$;
\ENDIF
\ENDFOR
\RETURN Adversarial image $I^*$.

\end{algorithmic}
\end{algorithm}


\subsubsection{Latent pairwise diffusion optimization} 

Gradient guidance and merging of latent variables are key steps to ensure that the representation of latent spaces during generation accurately guides the optimization of the generator $G$. Specifically, the gradient of the target model $f$ needs to be used first to guide the latent variable pairs:
\begin{equation}
    \begin{aligned}
    x_{t-1}^{\mathrm{inter}}=x_{t}-\eta_{x}\nabla_{x_{t}}\ell(x_{t},y_{target});\\
    y_{t-1}^{\mathrm{inter}}=y_{t}-\eta_{y}\nabla_{y_{t}}\ell(y_{t},y_{target}),
\end{aligned}
\label{eq9}
\end{equation}
where $L$ denotes the loss function, $\nabla$ indicates the gradient of $f$, $y_{target}$ indicates the attacker specifies the target label of the attack, and $\eta$ denotes the learning rate of the latent variable pairs $(x_{t},y_t)$.

After the gradient-guided optimization, we merge the two optimized latent variable pairs $(x_{t-1},y_{t-1})$ into a single unified latent variable $z_{t-1}$ to continue the diffusion model generation process:
\begin{align}
\label{eq10}
\begin{gathered}
    z_{t-1}=\alpha\cdot x_{t-1}^{\mathrm{inter}}+(1-\alpha)\cdot y_{t-1}^{\mathrm{inter}}, 
\end{gathered}
\end{align}
where $\alpha\in[0,1]$ is a weight parameter to control the ratio of the contribution of the two latent variables in the merger. The goal of the merging process is to combine the optimized properties of the two latent spaces, producing a more robust and integrated latent representation that better supports the generation of diffusion models.

\subsubsection{Adversarial example synthesis}

In DAC, $P^{\prime}$ is provided to the generator G as the input and initial condition for the adversarial example generation process. During the reverse diffusion process of the generator, we leverage EDICT to design parameter-shared coupled noise variable pairs $(x_t,y_t)$ in the diffusion model generation process and achieved robust inversion through an iterative bidirectional calibration mechanism at discrete time steps. In each diffusion step, one of the variables calculates the denoising direction based on the state of the other variable, and the result of this calculation is immediately incorporated into the conditional update input of the other vector. This cross-feedback mechanism constructs a dynamic error correction system. This method approximates the initial input by performing one denoising operation at each time step, enabling the model gradient to propagate backward through the entire chain. At each step, gradient guidance is utilized to progressively refine the generated image toward the target class $y_{target}$. Specifically, for each latent variable $z_t$ in the generation process, joint optimization of $(x_t,y_t)$ is performed, leveraging the gradient information of the target loss to guide the latent variable closer to the target class during optimization. At each time step $t$, DAC adjusts the variable pair $(x_t,y_t)$ through gradient guidance, controlling the diffusion and denoising processes. The synergy between gradient guidance and latent variable optimization ensures that the generated adversarial examples not only precisely align with the target conditions, but also achieve optimal performance in terms of distribution consistency and attack efficacy. Our adversarial example generation can be mathematically expressed as:
\begin{align}
    z'=\arg\min_{z_t}\ell(z_t,P,y_{target}),
\end{align}
where $z'$ denotes the final latent variable. After this, we pass $z'$ to the generator $G$ to obtain the final output of the adversarial examples $I^* = G(z')$. The specific adversarial example synthesis process is summarized in Algorithm \ref{alg:adversarial example synthesis}.




\subsection{Bi-path optimization strategy}



Although DAC can effectively deceive the machine into producing false recognition results by utilizing the original model information in a white-box environment, the premise of a white-box environment is that the attacker has access to the complete structure and parameters of the target model. In practical applications, especially in adversarial CAPTCHA scenarios, this premise often does not hold because the detailed information of the target model is not available, and direct access to the original image source may not be available due to the copyright of the image source. Therefore, we extend the DAC method into a more practical black-box attack method, called BP-DAC.

In BP-DAC, prior to the attack phase, we leverage the CLIP model's gradient \cite{radford2021learningtransferablevisualmodels} to guide generator G in producing the initial states $x_m$ and $y_m$; following CLIP's bootstrap optimization, these states are synthesized via Eq.(\ref{eq10}) into a temporary latent variable $z_m$. This CLIP-guided initialization critically enforces consistency between the generated image and the target distribution $P^{\prime}$, enhancing naturalness, visual quality, and preventing distribution drift, while significantly bolstering adversarial stealth through alignment with the prompt description, thus evading both human and automated detection. During the attack phase, we integrate gradients from multiple models and introduce a bi-path optimization strategy. This strategy addresses white-box limitations and enhances black-box adaptability by approximating the target model's behavior through merged predictions, thereby boosting the Attack Success Rate (ASR) on unknown models. The bi-path component specifically balances losses derived from the target class and the second-highest posterior probabilities, effectively exploiting vulnerabilities at the decision boundary to jointly enhance adversarial robustness and ASR.


Then, we integrate the gradients of multiple models during the attack and propose a bi-path optimization strategy to address the limitations of white-box environments and adapt to black-box scenarios. By merging predictions from multiple models, we can approximate the behavior of the target model, enhancing the ASR on unknown models. The bi-path strategy balances the loss from the target class and the second-highest posterior probabilities in the target model's output, effectively capturing vulnerabilities in the decision boundary to improve adversarial robustness and ASR.

Compared to traditional attacks, BP-DAC excels in generating perturbations by exploring the input space more comprehensively through a bi-path optimization strategy. This increases the diversity of adversarial examples and enhances the attack's robustness, allowing it to effectively exploit weaknesses in the target model. As a result, BP-DAC achieves higher success rates in complex and uncertain black-box environments. 


To accomplish the above, we need to add the assumption that we have three known proxy models $f_\mathrm{1}$,$f_\mathrm{2}$ and $f_\mathrm{3}$, which have classification predicted probability distribution of $g_1(x)$, $g_2(x)$, $g_3(x)$, and the parameters $\beta_1$, $\beta_2$ and $\beta_3$, respectively. Using the predict probability distribution of these models, we update the latent variables $x_t$ and $y_t$. Based on DAC, we replace the known model gradients in Eq.(\ref{eq9}) and Eq.(\ref{eq10}) with the weighted average of the gradients from the three proxy models to enhance the attack's effectiveness against an unknown black-box model. Specifically, our integrated classification predicted probability distribution $g_{x_t}$ and $g_{y_t}$ can be expressed as:
\begin{equation}
    \begin{aligned}
    g_1(x)=\mathrm{u}_1=\left[u_1^{(1)}, u_1^{(2)}, \ldots, u_1^{(M)}\right]^T \\
    g_2(x)=\mathrm{u}_2=\left[u_2^{(1)}, u_2^{(2)}, \ldots, u_2^{(M)}\right]^T \\
    g_3(x)=\mathrm{u}_3=\left[u_3^{(1)}, u_3^{(2)}, \ldots, u_3^{(M)}\right]^T \\
    g_{x_t} = \frac{\beta_{1}g_1(x) + \beta_{2}g_2(x) + \beta_{3g_3(x})}{\beta_{1}+\beta_{2}+\beta_{3}},
\end{aligned}
\end{equation}
where $g_1,g_2,g_3: \mathbb{R}^{H \times W \times C} \rightarrow \mathbb{R}^M$, and the calculation of $g_{y_t}$ is the same as for $g_{x_t}$.

\begin{algorithm}[t]
\caption{Adversarial example synthesis in BP-DAC}
\label{alg:bpDAC}
\begin{algorithmic}[1]
\REQUIRE Initial prompt $P$ from the attacker, clip model $M$, origin category $y_{origin}$, learning rates $\eta_x$, $\eta_y$, proxy models $f_\mathrm{1}$, $f_\mathrm{2}$ and $f_\mathrm{3}$, target model $f_t$
\ENSURE $f(I_t) \neq y_{origin}$;
\STATE $P' = f_{LLM}(P)$;
\STATE Initialize latent variables $x$ and $y$
\STATE Using Eq. (9) and Eq. (10) optimize the latent variables $x$ and $y$ through $m$, get $x_m$ and $y_m$
\FOR{$t = T$ to $1$}
\STATE $I_{t}=G(z_t,P^{\prime},\epsilon_t)$;
\IF {$f_t(I_{t}) \neq y_{origin}$}
\STATE exit the loop and proceed to Step 20
\ELSE
\STATE Compute the probability distributions $g_1(x)$, $g_2(x)$, $g_3(x)$ of the proxy models
\STATE Integrate $g_1(x)$, $g_2(x)$, $g_3(x)$ as $g_{x_t}$ base on Eq. (12)
\STATE Compute loss of second-highest and target class $\ell_{div}$, $\ell_{tar}$ 
\STATE Integrate $\ell_{div}$, $\ell_{tar}$ as $\ell_{x_t}$ base on Eq. (14)
\STATE Compute gradient $\nabla$ base on Eq. (15)
\STATE Update latent variables $x$ and $y$:
\STATE $x^{inter}_{t_1} = x - \eta_x * \nabla_{x_t} \ell_{x_t}$; 
\STATE $y^{inter}_{t-1} = y - \eta_y * \nabla_{y_t} \ell_{y_t}$;
\STATE Merge optimized latent variables into a unified representation:\\
$z_{t-1} = \alpha * x^{inter}_{t-1} + (1 - \alpha) * y^{inter}_{t-1}$;
\STATE $I_t = G(z_{t-1})$;
\ENDIF
\ENDFOR
\RETURN Adversarial image $I^*$.

\end{algorithmic}
\end{algorithm}

Leveraging the collective gradient information from these proxy models, we further guide the model to generate adversarial examples toward the second-highest and target class probabilities. The bi-path optimization strategy identifies a more robust path across different loss spaces, enhancing the transferability and success rate of adversarial examples in black-box models. This approach effectively improves the performance of black-box attacks, making adversarial examples more deceptive and robust.
 
Specifically, we focus on guiding the generation process by minimizing the loss associated with both the second-highest and target class probabilities:
\begin{equation}
    \begin{aligned}
    \ell_{div} &= -\ell(x_t,y_{second}); \\
    \ell_{tar} &= \ell(x_t,y_{target}).
\end{aligned}   
\end{equation}


Finally, the new loss $\ell_{BP}$ is derived by aggregating the losses from multiple directional objectives, effectively balancing the guidance effects of both the target and auxiliary classes during optimization. This overcomes the challenges associated with black-box model attacks, as given by:
\begin{equation}
    \begin{aligned}
        \ell_{x_t} &= \frac{\ell_{div} + \ell_{tar}} {2}.
    \end{aligned}
\end{equation}

Compute gradient information through loss $\ell_{BP}$:
\begin{equation}
    \begin{aligned}
        \nabla_{x_t} = \frac{\partial\ell_{x_t}}{\partial g{x_t}}.
    \end{aligned}
\end{equation}

The specific adversarial example synthesis process is shown in Algorithm \ref{alg:bpDAC}.

\begin{table}[t]
\caption{{Some of the prompt for LLM and diffusion model.}}
\label{tab:metrics}

\centering
\begin{tabular}{c|c}
\toprule[0.15em]

Origin Class &Input Prompt \\ \midrule 
Tench & A tench is swimming in the pond \\ \midrule 
Goldfish & A goldfish swims in the bowl \\  \midrule 
White shark & A white shark is hunting \\ \midrule 
Hummingbird & A hummingbird hovers near the flower \\  \midrule 
White stork & A white stork stands by the river \\ \midrule 
Marmot & A marmot stands on a rock   \\  \midrule 
Otter & An otter swims in the river \\ \midrule 
Gila monster & A Gila monster hides under a rock  \\   \midrule 
School bus&A school bus carries the children \\  \midrule 
Toilet paper&Toilet paper is rolled on the holder\\  

\midrule [0.15em]
\end{tabular}

\label{tab:pitch}
\end{table}

\begin{table*}[t]
\caption{Targeted attack success rates (\%) against black-box target models with the four source models. For each attack, we also reported the average attack success rate. The best results are in \textcolor{blue}{blue}. $^*$ indicates that surrogate model and target model are same.}
\label{tab:asr_table}
\resizebox{1\textwidth}{!}{%
\begin{tabular}{lcccccccccc}
\toprule[0.15em]
\textbf{Target} & \multicolumn{9}{c}{Method} &  \\
\cmidrule(l){2-10}

\textbf{Model} & SAE & ADer & ReColorADV & cAdv & tADV & NCF & ACE & ColorFool & ACA & DAC  \\
\midrule

RN-50&88.0$^*$&55.7$^*$&96.4$^*$&97.2$^*$&99.0$^*$&\textcolor{blue}{99.1$^*$}&90.1$^*$&91.4$^*$&88.3$^*$&98.1$^*$\\
RN-152&46.5& 7.8& 33.3& 37.0& 30.2& 15.2& 21.0& 60.5& 61.7&\textcolor{blue}{79.9}\\
MN-v2&63.2& 15.5& 40.6& 44.2& 43.4& 32.8& 41.6& 71.2& 69.3&\textcolor{blue}{89.3}\\
Dense-161&41.9&8.4&28.3&36.8&28.8&16.1&18.6&48.5&61.9&\textcolor{blue}{87.9}\\
Eff-b7&28.8&11.4&19.2&34.9&21.6&12.7&15.4&32.4&60.3&\textcolor{blue}{61.3}\\
Inc-v3&25.9&7.7&17.7&25.3&27.0&9.4&9.8&33.6&61.6&\textcolor{blue}{75.0}\\
\cdashline{1-11}
\rowcolor{gray!20}Average&49.05&17.75&39.25&45.9&41.7&30.9&32.75&56.3&67.2&\textcolor{blue}{81.9}\\

\midrule[0.1em]

\textbf{Target} & \multicolumn{9}{c}{Method} &  \\
\cmidrule(l){2-10}
\textbf{Model}&SAE&ADer&ReColorADV&cAdv&tAdv&NCF&ACE&ColorFool&ACA&DAC\\
\midrule
RN-50&53.2&8.4&33.7&39.6&31.5&17.9&25.7&65.9&62.6&\textcolor{blue}{88.9}\\
RN-152&41.9&7.1&26.4&29.9&24.5&12.6&15.4&56.3&56.0&\textcolor{blue}{79.6}\\
MN-v2&90.8$^*$&56.6$^*$&97.7$^*$&96.6$^*$&\textcolor{blue}{99.9$^*$}&99.1$^*$&93.3$^*$&93.2$^*$&93.1$^*$&91.6$^*$\\
Dense-161&38.0&7.7&24.7&33.9&24.3&12.4&15.3&43.5&55.7&\textcolor{blue}{84.4}\\
Eff-b7&26.9&10.9&20.7&32.7&22.4&11.7&13.4&33.0&51.0&\textcolor{blue}{69.5}\\
Inc-v3&22.5&7.6&18.6&26.8&27.2&9.5&9.5&33.6&56.8&\textcolor{blue}{64.5}\\
\cdashline{1-11}
\rowcolor{gray!20}Average&45.55&16.4&37.0&43.25&38.3&27.2&28.8&54.25&62.5&\textcolor{blue}{79.75}\\

\bottomrule
\end{tabular}%
}
\label{tab:dac_compare}
\end{table*}

\begin{table*}[t]
\caption{The performance comparison of UAC and some traditional white-box adversarial attack on target attack success rate (TSR) and untarget attack success rate (USR).}

\footnotesize
\resizebox{\linewidth}{!}{%
\begin{tabular}{c|c|ccccccc}

\midrule[0.15em]
\multicolumn{2}{c|}{\multirow{3}*{Attack Methods}}&\multicolumn{7}{c}{\textbf{TSR Label} : Gila Monster} 
\\\cmidrule(l){3-9}
\multicolumn{2}{c|}{~}&{ResNet50}&{ResNet152}&{SeResNext101}&{Effecientnet}&{Googlenet}&{MobileNetV2}&{Alexnet} 
\\\midrule
\multirow{2}{*}{PGD}&{ USR }&{99.8\%}&{100\%}&{99.8\%}&{99.9\%}&{99.6\%}&{99.7\%}&{99.9\%} 
\\\cmidrule(l){2-9}
&{ TSR }&{69.8\%}&{65.3\%}&{47.7\%}&{46.8\%}&{77.3\%}&{81.1\%}&{65.0\%} 
\\\midrule
\multirow{2}{*}{FGSM}&{ USR }&{100\%}&{99.7\%}&{100\%}&{99.9\%}&{100\%}&{100\%}&{100\%} 
\\\cmidrule(l){2-9}
&{ TSR }&{75.3\%}&{60.0\%}&{62.3\%}&{53.6\%}&{81.7\%}&{83.5\%}&{64.3\%} 
\\\midrule
\multirow{2}{*}{BIM}&{ USR }&{100\%}&{100\%}&{100\%}&{99.9\%}&{100\%}&{100\%}&{100\%} 
\\\cmidrule(l){2-9}
&{ TSR }&{75.3\%}&{67.9\%}&{62.3\%}&{53.6\%}&{81.7\%}&{83.5\%}&{64.3\%} 
\\\midrule
\multirow{2}{*}{DAC}&{ USR }&{100\%}&{100\%}&{100\%}&{100\%}&{100\%}&{100\%}&{100\%} 
\\\cmidrule(l){2-9}
&{ TSR }&{100\%}&{100\%}&{100\%}&{100\%}&{\textbf{99\%}}&{100\%}&{\textbf{99\%}} 
\\

\bottomrule
\end{tabular}
}

\label{tab:white}
\end{table*}
\begin{table*}[t]
\caption{The performance comparison of BP-DAC and some state-of-art adversarial attack methods relying on generative models on ASR.}

\footnotesize
\resizebox{\linewidth}{!}{%
\begin{threeparttable}
\begin{tabular}{ccc|ccccccc}

\midrule[0.15em]
\multicolumn{3}{c|}{\multirow{3}*{Baseline Models}}&\multicolumn{7}{c}{\textbf{Attack method} : BP-DAC} 

\\\cmidrule(l){4-10}
\multicolumn{3}{c|}{~}&{ResNet50}&{ResNet152}&{SeResNext101}&{Effecientnet}&{Googlenet}&{MobileNetV2}&{Alexnet} 
\\\midrule
{ResNet50}&{SeResNext101}&{Googlenet}&{100\%}&{98.5\%}&{100\%}&{99.1\%}&{100\%}&{97.2\%}&{  98.0\%} 
\\\midrule
{ResNet50}&{Effecientnet}&{Alexnet}&{100\%}&{98.0\%}&{98.0\%}&{100\%}&{97.6\%}&{97.9\%}&{100\%} 
\\\midrule
{Googlenet}&{Effecientnet}&{Alexnet}&{97.7\%}&{96.6\%}&{97.2\%}&{100\%}&{100\%}&{95.5\%}&{100\%} 
\\\midrule
{ResNet152}&{MobileNetV2}&{Alexnet}&{95.4\%}&{100\%}&{95.8\%}&{96.2\%}&{97.2\%}&{100\%}&{100\%} 
\\\midrule
{ResNet152}&{SeResNext101}&{MobileNetV2}&{99.3\%}&{100\%}&{100\%}&{99.4\%}&{99.2\%}&{100\%}&{99.2\%} 
\\\bottomrule

\end{tabular}

\end{threeparttable}
}

\label{tab:black}
\end{table*}

\section{Experiments and Results}
\label{sec:3}


In this section, we first validate the effectiveness of our method in white-box and black-box scenarios by comparing it with other methods and calculating the attack effectiveness of existing methods in the face of unknown models. Next, we conduct a comprehensive ablation experiment to compare the attack effect, the quality of the generated images, and the efficiency of the attack, thereby proving the usefulness of our module and the rationality of our parameter settings.


\subsection{Experimental setup}
\subsubsection{Dataset}
Imagenet is one of the most widely used datasets for image classification, object detection, and object localization in deep learning. Although this experiment did not require the images provided by the dataset as the basis for the input, we refer to Imagenet with 1000 classifications as input and attacked categories in this experiment.

\subsubsection{Environment} 
All experiments were conducted on an Ubuntu 22.04.4 Server with an Intel(R) Xeon(R) Gold 6342 CPU @ 2.80 GHz and an NVIDIA A40 with 48 GB of memory.

\subsubsection{Attack Model}


All tasks in this experiment were performed based on eight models, including SeResNeXt101 \cite{xie2017aggregatedresidualtransformationsdeep}, MobilenetV2 \cite{sandler2018mobilenetv2}, Resnet50 \cite{he2015deepresiduallearningimage}, Resnet152, Googlenet \cite{szegedy2014goingdeeperconvolutions}, Efficientnet \cite{tan2020efficientnetrethinkingmodelscaling}, inceptionV3 \cite{szegedy2015rethinkinginceptionarchitecturecomputer}, and Alexnet \cite{2012ImageNet}.

\subsubsection{Evaluation Metrics}

To demonstrate the effectiveness and superiority of our sparse attack method, we employ the Attack Success Rate (ASR), the Clip Score \cite{hessel2022clipscorereferencefreeevaluationmetric}, and the Average Attack Step compared to other methods.

ASR measures the proportion of inputs successfully manipulated to produce the attacker's desired erroneous outputs.
\begin{equation}
    \mathrm{ASR}=\frac{N_{adv}}{N_{total}}\times100\%,
\end{equation}
where $N_{adv}$ is the number of adversarial examples that successfully mislead the target model, while $N_{total}$ denotes the total number of generated adversarial examples.

Clip Score evaluates semantic alignment between images and text using the Clip model, commonly applied in tasks like image generation and text-to-image synthesis:
\begin{equation}
    \text{Clip Score(I, C)}=max(100*cos(E_I,E_C),0),
\end{equation}
where $I$ is the input image, $C$ is the input text description, and $cos(E_I,E_C)$ denotes the cosine similarity between the image vector $E_I$ and the text vector $E_C$.

\subsection{Comparison and analysis of attack effects}

To demonstrate the superior performance of our method in both white-box and black-box scenarios, we conducted comprehensive comparative experiments with DAC and Bi-DAC in both white-box and black-box scenarios.

As shown in Table \ref{tab:dac_compare}, we compare the ASR of ten adversarial attack methods (e.g., SAE \cite{hosseini2018semanticadversarialexamples}, Adef \cite{alaifari2019adefiterativealgorithmconstruct}, ReColorAdv \cite{laidlaw2019functionaladversarialattacks}, cAdv \cite{bhattad2020unrestrictedadversarialexamplessemantic}, tAdv \cite{bhattad2020unrestrictedadversarialexamplessemantic}, ACD \cite{zhao2020adversarialcolorenhancementgenerating}, ColorFool \cite{shamsabadi2020colorfoolsemanticadversarialcolorization}, NCF \cite{yuan2022naturalcolorfoolboosting}, ACA \cite{chen2023contentbasedunrestrictedadversarialattack} and DAC) which craft adversarial examples not require the addition of perturbation on seven deep learning models. The experimental results demonstrate that DAC achieves remarkable performance in the black-box adversarial attack task. First, in the face of an unknown model, when the alternative model is different from the target model, the attack success rate of our method far exceeds that of other methods, and is only slightly lower than that of NCF when the alternative model is the same as the target model, which reflects the excellent migratory nature of our method; second, the average attack success rate of our method under two alternative models, RN-50 and MN-v2, reaches 81.9\% and 79.75\%, which is an improvement of more than 14\% over the suboptimal methods, and a significant lead in comprehensive performance; moreover, for complex models, our method shows stronger adaptability, and the advantage is especially prominent in lightweight or high-complexity models, which further proving the superiority of the defensive verification codes generated by our method when facing unknown models.

As shown in Table \ref{tab:white}, we compare the ASR of four classical adversarial attack methods (e.g., PGD \cite{madry2019deeplearningmodelsresistant}, FGSM \cite{goodfellow2015explainingharnessingadversarialexamples}, BIM \cite{kurakin2017adversarialmachinelearningscale}, and DAC) on seven deep learning models, including the untargeted attack success rate (USR) and the targeted attack success rate (TSR), with the target category “Gila Monster”. The experimental results show that the DAC method exhibits a nearly 100\%  USR and TSR in all models and scenarios. It significantly outperforms the other attack methods in both USR and TSR, demonstrating its overall robustness and attack effectiveness advantages. In contrast, although the other methods also achieve a high success rate (more than 99\%) in untargeted attack scenarios, the ASR is significantly lower in targeted scenarios, especially for the Efficientnet and SeResNext101 models, which suggests that there is still a large room for optimization of the performance of the traditional adversarial attack methods in targeted attacks. Untargeted attacks are highly mature under existing methods, and almost all methods can achieve a very high USR in white-box untargeted attack scenarios. However, the effectiveness of targeted attacks is limited by the complexity of the method design and model architecture. The DAC methods perform well in both targeted and untargeted attack scenarios, show good generalization and robustness, and provide a strong benchmark for subsequent research.

\begin{figure}
\centering
  \includegraphics[width=1\linewidth]{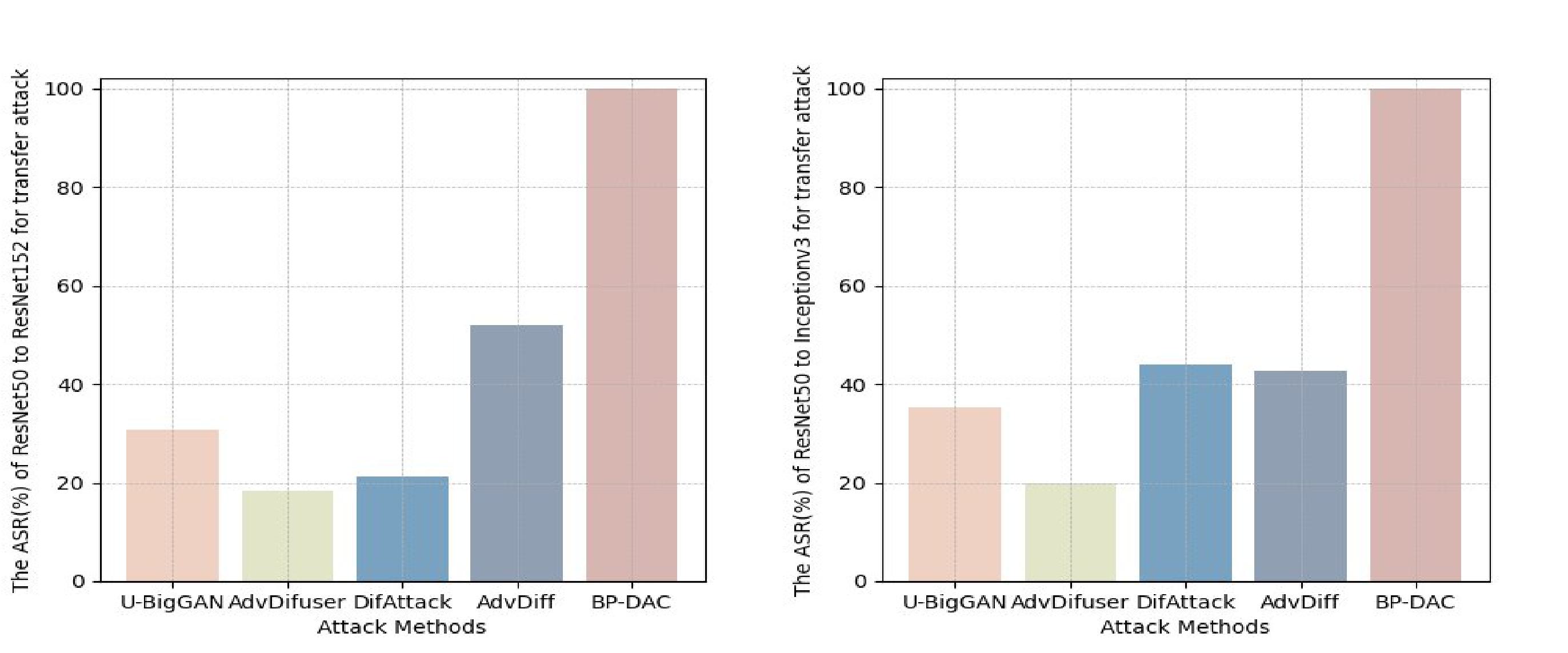}
  \caption{Attack success rate of transfer attacks based on Resnet50 (left) and InceptionV3 (right).}
  \label{compare}
\end{figure}

\begin{figure*}
    \centering
    \includegraphics[width=1\linewidth]{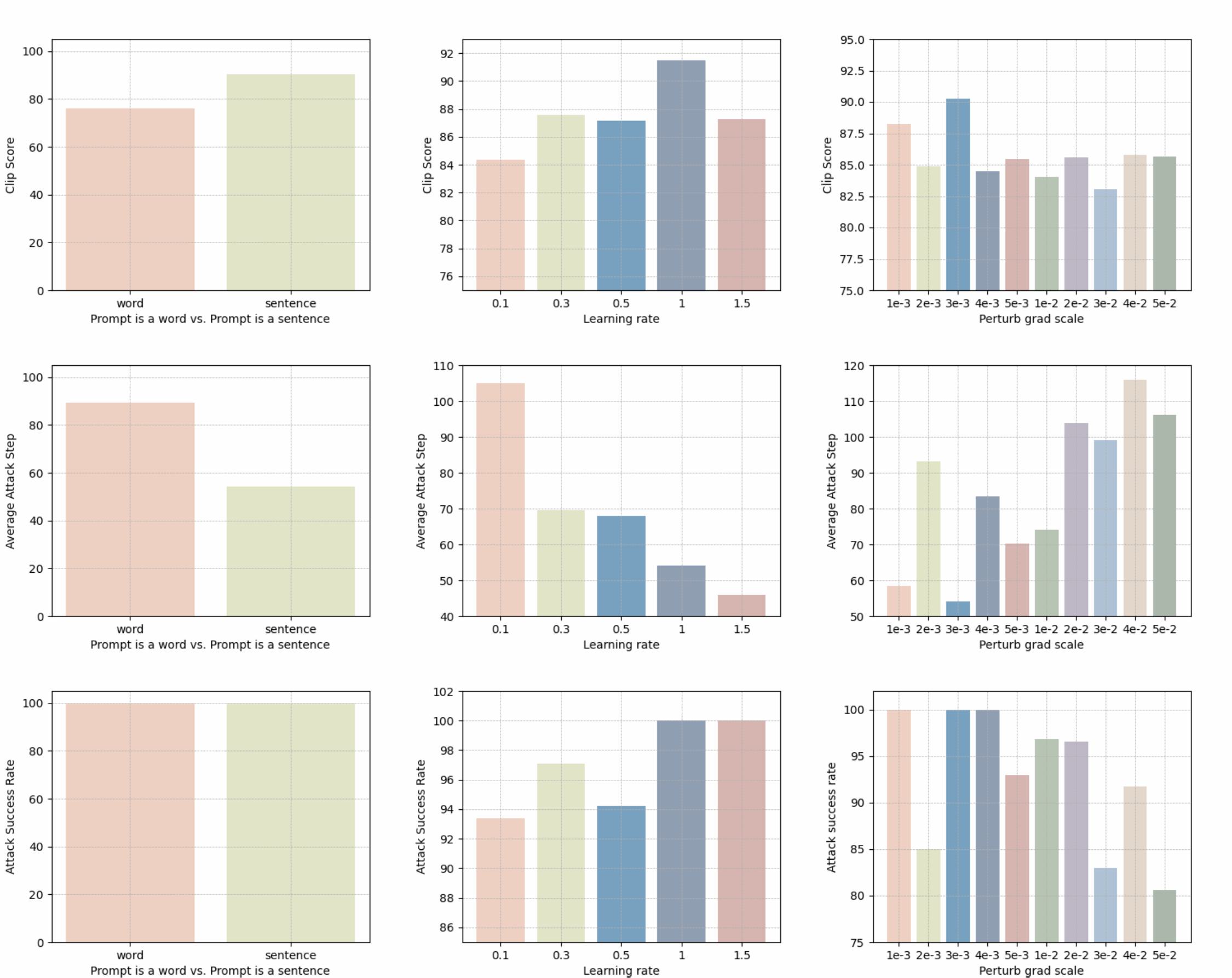}
    \caption{Attack success rate, average attack steps, and Clip Score of different settings.}
    \label{fig:ablation}
\end{figure*}

As shown in Table \ref{tab:black}, we evaluate the ASR using the BP-DAC method on different combinations of baseline and targeted attack models. Overall, the migration ASR of the BP-DAC method is higher than 99\% on all model combinations, regardless of the architectural differences between the baseline and target models, which demonstrates the strong generalization ability and robustness of the BP-DAC method in cross-model attacks. On the models of ResNet series, Googlenet, and MobileNetV2, the success rates of the attacks are almost indistinguishable whether they are the baseline model or the target model, which demonstrates that the antiperturbation generated by the BP-DAC method is highly adaptable and stable. It is worth noting that even for models with relatively complex architectures (e.g., SeResNext101 and Efficientnet), the success rate of BP-DAC's transfer attack remains exceptionally high, further demonstrating its performance advantage in dealing with diverse deep learning models. The experimental results validate the efficiency and robustness of BP-DAC in cross-model scenarios, reflecting its potential as a generalized adversarial attack method and providing critical experimental benchmarks and theoretical support for subsequent research.


As shown in Fig. \ref{compare}, we use the BP-DAC method to compare the transfer attack performance with four adversarial attack methods based on generative models using ResNet50 as the baseline model on ResNet152 and InceptionV3 models. It can be observed that existing adversarial attack methods based on the diffusion model exhibit poor attack performance against unknown models, and only the AdvDiff method can barely exceed a 50\% ASR. In contrast, our method still achieves a success rate of nearly 100\% when facing an unknown model. This further demonstrates the superiority of our method in the face of the unknown model. In the production of CAPTCHA, since the illegal model used by the attacker is unknown, the method for adversarial CAPTCHA defense must have a high degree of robustness in the face of the unknown model.





\subsection{Ablation study}


\begin{figure*}[t]
\centering
  \includegraphics[width=1\linewidth]{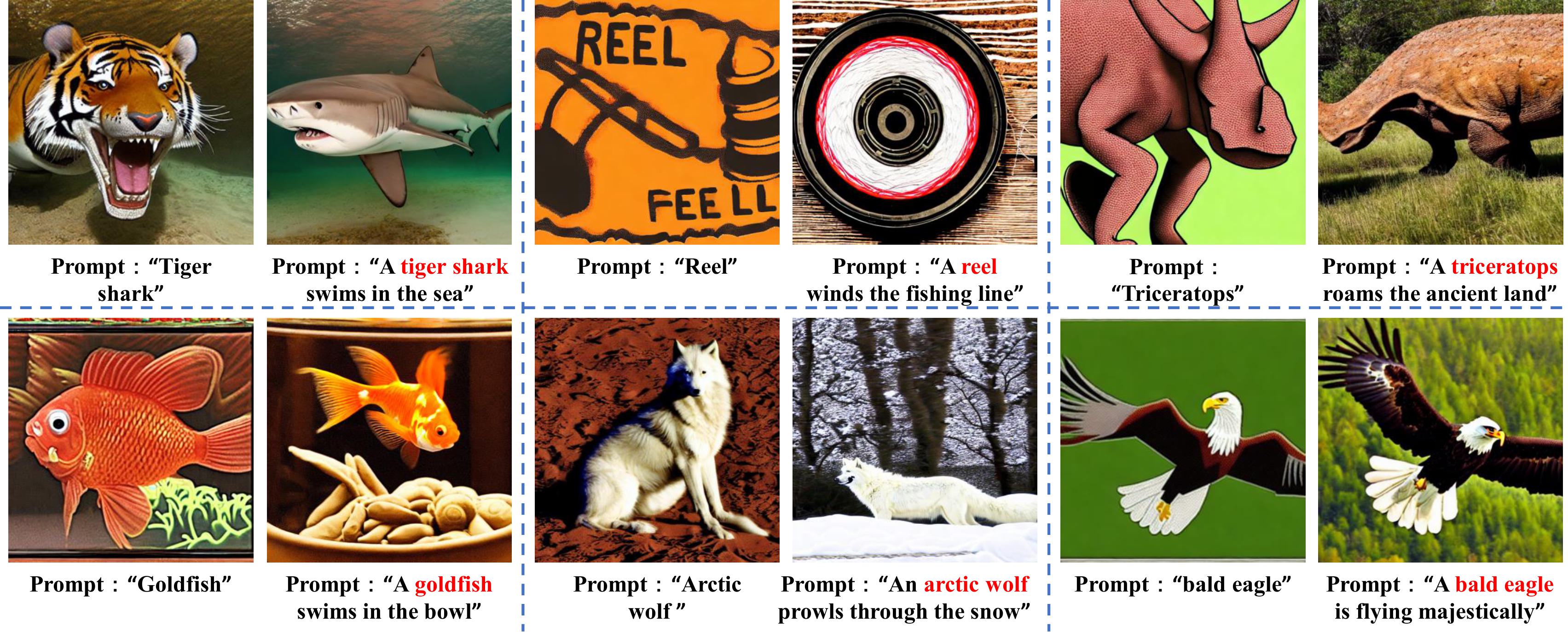}
  \caption{Comparison plot of adversarial samples generated using detailed prompt versus using labels. It is easy to conclude that a more detailed prompt will result in a more accurate and higher-quality generated image.}
  \label{llm}
\end{figure*} 

We compare adversarial examples generated using class names from ImageNet as prompts with those generated using sentences as prompts, and incorporate Clip model gradients for optimization. As shown in the first column of Fig. \ref{fig:ablation}, although both types of prompts achieved an ASR of 100\%, the average number of steps required for a successful attack is noticeably higher when the prompt is a single word. Additionally, the Clip Score of the generated images is lower, indicating poorer quality of the adversarial examples. By incorporating clip model gradients during the generation process, we ensure semantic consistency and visual alignment, which further improves the quality of the generated examples. Overall, the combination of prompt optimization through LLM and gradient optimization via the CLIP model significantly enhances both attack efficiency and the quality of adversarial examples. Furthermore, as demonstrated in Fig. \ref{llm}, when the prompt consists of a single word, the simplicity and lack of information in the semantic expression lead to the generation of incorrect or unrealistic images. In contrast, our optimization strategy effectively guides the generation model to output high-quality images that are better aligned with the target class. This comparison demonstrates the crucial role of both LLM and clip model in our method, as they not only improve the quality of the generated examples but also significantly increase the attack efficiency.


\begin{table}[h]
\caption{{The ASR of adversarial attacks when againsting different defence methods.}}
\label{tab:metrics}
\resizebox{\linewidth}{!}{
\centering
\begin{tabular}{c|cccc}
\toprule[0.15em]

\multirow{2}{*}{Attack Models}&\multicolumn{4}{c}{\textbf{Baseline Model} : ResNet50 } \\\cmidrule(l){2-5}&NRP &R\&P&RS&HGD\\\midrule
U-BigGAN&30.9\%&14.2\%&34.5\%&22.6\%  \\  
AdvDiffuser&40.5\%&15.4\%&38.4\%&10.8\%  \\  
DiffAttack&38.5\%&23.7\%&40.8\%&20.5\%  \\  
AdvDiff&74.2\%&56.8\%&82.8\%&53.8\%  \\  
BP-DAC&100\%&97.1\%&100\%&95.7\%  \\ 

\midrule [0.15em]
\end{tabular}
}
 
\label{tab:defence}
\end{table}


\subsection{Robustness evaluation}

In traditional adversarial attack scenarios, various defense preprocessing methods are commonly employed to reduce the efficacy of adversarial samples. However, in practical applications of adversarial CAPTCHA systems, if an attacker preprocesses the CAPTCHA using these defense mechanisms during the attack process, the security of the CAPTCHA is significantly compromised. To validate the robustness of our proposed adversarial CAPTCHA system, this study extensively analyzes the impact of various defense strategies, including NRP \cite{naseer2020selfsupervisedapproachadversarialrobustness}, RS \cite{cohen2019certifiedadversarialrobustnessrandomized}, R\&P \cite{xie2018mitigatingadversarialeffectsrandomization}, and HGD \cite{liao2018defenseadversarialattacksusing}, on adversarial samples. As illustrated in Table \ref{tab:defence}, the potency of attack methods based on generative models is typically diminished after defense preprocessing, likely due to increased uncertainty in the samples, which reduces the effectiveness of attacks based on generative models. Nonetheless, our BP-DAC, through a bi-path optimization strategy, overcomes the limitations of traditional attack approaches and effectively mitigates the impact of defense measures. By integrating gradient information from multiple models, BP-DAC enhances the resilience of adversarial CAPTCHAs against various defense mechanisms, while mitigating the issue of model overfitting.




\subsection{Influence and setting of Parameters}

We also conducted a comprehensive ablation experiment to verify the parameter settings. We calculated the Clip Score, average attack step and ASR of the adversarial examples generated with different learning rate and different perturb grad scale, as show in the second and third columns of Fig. \ref{fig:ablation}, with the increase of the learning rate, the average attack step required for the success of the attack decreases significantly. However, a higher learning rate $\eta$ can cause the generative model to be over-guided in some steps during the generation process, thus decreasing the quality of the generated image. For this reason, we consider the quality of the image, the efficiency of the attack, and choose to set $\eta$ to 1.0. At the same time, for the size of the perturbation $\epsilon$, in order to guarantee the method is effective, we must ensure the success rate of the attack. In the case of ASR of 100\%, we choose to set the perturb grad scale to $3\times10^{-3}$ that requires the least number of attack steps and generates the highest Clip Score of the adversarial examples.

\section{conclusion}


In this paper, we proposed a novel defensive adversarial CAPTCHA generation framework, BP-DAC, which integrates LLM and generative models while introducing an innovative bi-path adversarial optimization strategy. By overcoming the limitations of traditional adversarial attack methods that add noise to original images, BP-DAC leverages gradients from multiple deep models and simultaneously guides the model toward generating adversarial examples in the directions of the second-highest and target class probabilities. This approach enables BP-DAC to identify more robust paths in different loss spaces, achieving exceptionally high ASR even against unknown models. Experimental results demonstrated that our method not only generates realistic images but also effectively deceives traditional DNN recognition models in various white-box and black-box application scenarios, providing a new direction for future adversarial attack research. Furthermore, given the widespread use of CAPTCHAs in daily life for identity verification and security protection, our method can effectively enhance the security of CAPTCHA systems, preventing unauthorized intrusions and providing essential technical support for CAPTCHA design and security upgrades. In the future, we plan to further explore methods for generating higher-quality images.


\bibliographystyle{IEEEtran}
\bibliography{references}

\vfill

\end{document}